\documentclass[pdflatex,sn-mathphys-num]{sn-jnl}

\usepackage{graphicx}%
\usepackage{multirow}%
\usepackage{amsmath,amssymb,amsfonts}%
\usepackage{amsthm}%
\usepackage{mathrsfs}%
\usepackage[title]{appendix}%
\usepackage{xcolor}%
\usepackage{textcomp}%
\usepackage{manyfoot}%
\usepackage{booktabs}%
\usepackage{algorithm}%
\usepackage{algorithmicx}%
\usepackage{algpseudocode}%
\usepackage{listings}%
\usepackage{pgfplots}
\pgfplotsset{compat=1.17}
\usepackage{tikz}

\begin{document}

\title[Semantic Reconstruction of Adversarial Plagiarism]{Semantic Reconstruction of Adversarial Plagiarism: A Context-Aware Framework for Detecting and Restoring “Tortured Phrases” in Scientific Literature}


\author[1]{\fnm{Agniva} \sur{Maiti}}\email{maitiagniva@gmail.com}

\author[1]{\fnm{Prajwal} \sur{Panth}}\email{prajwal.panth21@gmail.com}

\author*[1]{\fnm{Suresh Chandra} \sur{Satapathy}}\email{suresh.satapathyfcs@kiit.ac.in}

\affil*[1]{\orgdiv{School of Computer Engineering}, \orgname{KIIT University}, \orgaddress{\city{Bhubaneswar}, \country{India}}}

\abstract{
The integrity and reliability of scientific literature is facing a serious threat by adversarial text generation techniques, specifically from the use of automated paraphrasing tools to mask plagiarism. These tools generate ``tortured phrases'', statistically improbable synonyms (e.g., ``counterfeit consciousness'' for ``artificial intelligence''), that preserve the local grammar while obscuring the original source. Most existing detection methods depend heavily on static blocklists or general-domain language models, which suffer from high false-negative rates for novel obfuscations and cannot determine the source of the plagiarized content. In this paper, we propose a novel, Semantic Reconstruction of Adversarial Plagiarism (SRAP), designed not only to detect these anomalies but to mathematically recover the original terminology. We use a two stage architecture: (1) Statistical Anomaly Detection using a domain-specific Masked Language Model (SciBERT) to compute token-level pseudo-perplexity, and (2) Source-Based Semantic Reconstruction utilizing dense vector retrieval (FAISS) and sentence-level alignment (SBERT). Experiments on a parallel corpus of adversarial scientific text show that while zero-shot baselines fail completely (0.00\% restoration accuracy), our retrieval-augmented approach achieves a restoration accuracy of 23.67\%, a significant improvement over baseline methods. Furthermore, we also show that static decision boundaries are necessary for robust detection in jargon-heavy scientific text, since dynamic thresholding fails under high variance. SRAP enables forensic analysis by linking obfuscated expressions back to their most probable source documents.}


\maketitle

\section{Introduction}\label{sec1}
The digitization of academic publishing has started an ``arms race'' between plagiarism detection systems and adversarial obfuscation techniques. Traditional plagiarism checkers, such as Turnitin or iThenticate, are mainly based on string matching and $n$-gram fingerprinting \cite{FoltynekEtAl2019}. Although effective against direct copy-pasting, these systems are easily bypassed by automated paraphrasing tools (spinbots) and Large Language Models (LLMs) prompted to rewrite content\cite{WahleEtAl2022}. A particularly pernicious artifact of this process is the ``tortured phrase'', a sequence of words that is grammatically correct but semantically unnatural within a specific scientific domain\cite{CabanacEtAl2021}. Common examples include ``colossal data'' instead of ``big data'' or ``malignant growth cell lines'' instead of ``cancer cell lines.''

The proliferation of such content is not just a nuisance, but a systemic threat to the integrity of the scientific record. As the pressure to ``publish or perish'' intensifies\cite{EdwardsRoy2017}, bad actors increasingly rely on generative AI to manufacture synthetic review papers or obfuscate stolen results. If left unchecked, these hallucinations risk polluting downstream meta-analyzes and corrupting the training datasets of future domain-specific language models\cite{ShumailovEtAl2023}. 

Current countermeasures are largely reactive and insufficient to address the scale of the problem. The ``Problematic Paper Screener'' maintains a static dictionary of known tortured phrases\cite{CabanacEtAl2021}, which is effective for retrospective analysis but is brittle against novel synonyms generated by advanced generative models. Conversely, probabilistic detectors (e.g., GPTZero) analyze text perplexity but operate as ``black boxes,'' offering a probability score without explanatory evidence or source attribution\cite{MitchellEtAl2023}. For forensic investigations, detecting an anomaly is insufficient; the investigator requires provenance, evidence of where the text originated, and what it initially stated.

\subsection{The Research Gap}
There is a critical lack of forensic tools capable of semantic reconstruction. Existing models treat tortured phrases as noise that should be flagged and discarded. We argue that these phrases are not random noise but mathematically aligned transformations of an original source, resulting from a specific vector shift in the embedding space. Therefore, it should be theoretically possible to reverse-engineer the obfuscation if the source context is retrievable.

However, this task presents significant challenges that general NLP approaches cannot address. Scientific literature is dense with jargon, making it difficult to distinguish between legitimate technical complexity and adversarial obfuscation\cite{LiangEtAl2023}. A standard English language model might interpret ``counterfeit consciousness'' as a poetic metaphor, whereas a domain-specific model must recognize it as a statistical aberration\cite{BeltagyEtAl2019}. Furthermore, standard ``dynamic thresholding'' techniques, which adapt to the average complexity of a document, often fail in this domain because an entire paper may be written in a tortured style, normalizing the anomaly score. Consequently, rigorous, empirically derived static thresholds are required to define a baseline for ``scientificness.''

\subsection{Contributions}
In this paper, we present the Semantic Reconstruction of Adversarial Plagiarism (SRAP) Framework, which combines statistical anomaly detection with retrieval-augmented generation (RAG) techniques to detect and restore tortured phrases. Our specific contributions are as follows:

\begin{description}
    \item[SciBERT-Based Anomaly Detection:] We employ a moving-window probability assessment system that uses \texttt{allenai/scibert\_scivocab\_uncased}\cite{BeltagyEtAl2019}. By computing the pseudo-perplexity scores of $n$-gram sequences against a language model trained on approximately 1.14 million peer-reviewed scientific documents, we derive a measurable quantification of linguistic irregularity. Through empirical sensitivity testing, we establish an optimal fixed anomaly score threshold of $-8.0$, calibrating the detection system to maximize true positives while mitigating spurious alerts common in domain-specific academic prose.
    
    \item[Source-Based Semantic Reconstruction:] We present an innovative term recovery approach that identifies candidate source documents from a reference collection using FAISS\cite{JohnsonEtAl2021} and performs cross-document sentence matching within vector representations using SBERT (\texttt{all-MiniLM-L6-v2})\cite{ReimersGurevych2019}. In contrast to unaided language models, which predict words based on syntactic context alone, our approach identifies and extracts replacement candidates directly from retrieved authentic source materials.
    
    \item[Forensic Validation:] We performed a systematic component analysis that compared our integrated approach with standalone baselines. The findings demonstrate that SciBERT-based detection alone cannot recover obscured terminology, yet incorporation of our semantic alignment and retrieval mechanism enables successful reconstruction of the original specialized vocabulary.
\end{description}

\section{Related Work}\label{sec2}

The detection of academic misconduct has evolved from surface-level string-matching to an adversarial race against increasingly capable paraphrasing and text-generation models.\cite{FoltynekEtAl2019,MitchellEtAl2023,KrishnaEtAl2023,WahleEtAl2022} This section situates the proposed \textbf{Semantic Reconstruction of Adversarial Plagiarism (SRAP)} framework within three critical bodies of literature: (1) the vulnerability of existing detectors to adversarial paraphrasing, (2) the necessity of domain-adaptive modeling in scientific forensics, and (3) the methodological shift from generative paraphrase inversion to retrieval-based reconstruction.\cite{KrishnaEtAl2023,BeltagyEtAl2019,xian2024bertenhancedretrievaltoolhomework,TangEfficientPD}

\subsection{The Vulnerability of Detectors to Adversarial Paraphrasing}

A central challenge in modern forensic text analysis is the robustness of detectors to paraphrasing attacks driven by models such as T5 variants and recent LLMs.\cite{KrishnaEtAl2023,WahleEtAl2022,AchiamEtAl2023} Early plagiarism and AI-text detectors relied heavily on 
n
n-gram overlap and simple stylometric fingerprints, but recent work shows that targeted paraphrasing can reduce detection accuracy for state-of-the-art detectors by large margins while preserving semantic content.\cite{MitchellEtAl2023,KrishnaEtAl2023} For instance, Krishna et al.\ demonstrate that a controlled paraphrasing pipeline can drop DetectGPT-style detectors toward near-random performance at fixed false-positive rates, revealing the fragility of current systems under paraphrase-based attacks.\cite{KrishnaEtAl2023,MitchellEtAl2023}

Within the scientific literature, this obfuscation often manifests itself as tortured phrases, such as bizarre rephrasing of common technical terms, which have been systematically cataloged by Cabanac et al. and operationalized in \emph{Problematic Paper Screener}.\cite{CabanacEtAl2021,cabanac2022problematicpaperscreenerautomatically} Screener and related tools rely on static lists of suspicious expressions and heuristic patterning, which are effective for retrospectively flagging known paraphrasing artifacts but remain brittle against open-ended synonymy and rephrasing produced by advanced LLMs. Recent surveys on AI-generated and LLM-generated text detection similarly highlight that black-box detectors and watermark-free methods are particularly vulnerable to paraphrasing, adversarial prompts, and humanization'' strategies that aim to mimic human style.\cite{WuEtAl2025, Wang2024Survey} This literature suggests that purely surface-level stylometry or fixed blocklists are insufficient for high-density scientific text subject to adaptive paraphrase attacks.\cite{KrishnaEtAl2023,WahleEtAl2022}

\subsection{Domain-Specific Anomaly Detection}

A major limitation of many forensic detectors is their lack of domain adaptation, especially for scientific corpora where domain-specific jargon and discourse structures can be mistaken for anomalous or ``unnatural'' text.\cite{WuEtAl2025,WahleEtAl2022,FoltynekEtAl2019} Work on scientific and technical language models such as SciBERT and MatSciBERT shows that pretraining or fine-tuning on scientific text substantially improves downstream performance on classification and extraction tasks, underscoring the importance of domain-aware representations.\cite{BeltagyEtAl2019,GuptaEtAl2022} Recent cross-domain detection studies similarly find that fine-tuned transformer-based detectors and perplexity-based measures calibrated on in-domain data achieve markedly better robustness than off-the-shelf general models.\cite{WuEtAl2025,Wang2024Survey}

Standard anomaly-detection pipelines often apply dynamic thresholding,'' normalizing scores relative to corpus- or document-level baselines, which can be problematic when an entire paper is consistently obfuscated or paraphrased.\cite{Fraser_2025,KrishnaEtAl2023} In such settings, the average weirdness'' baseline is itself shifted, potentially masking localized anomalies that would stand out under a globally calibrated criterion.\cite{WuEtAl2025,Wang2024Survey} Domain-specific language models trained on scientific corpora (e.g., SciBERT and MatSciBERT) provide a more appropriate reference distribution for perplexity and embedding-based measures, and static, empirically calibrated thresholds derived from such models can better capture fine-grained deviations in scientific discourse than generic language models.\cite{BeltagyEtAl2019,GuptaEtAl2022} The SRAP framework adopts this domain-adaptive perspective by basing anomaly scoring and candidate reconstruction on scientific embeddings rather than general-domain text statistics.\cite{BeltagyEtAl2019,xian2024bertenhancedretrievaltoolhomework,TangEfficientPD}

\subsection{Paraphrase Inversion vs.\ Semantic Retrieval}

Recent advances in handling paraphrased or obfuscated text fall into two broad camps: generative paraphrase inversion and retrieval-based reconstruction.\cite{KrishnaEtAl2023,gohsen-etal-2024-task,QiuEtAl2022} Generative approaches treat recovery of the original text as a conditional generation or translation problem, training models to map paraphrased inputs back to more canonical formulations, and have shown notable gains over baselines in preserving semantics.\cite{gohsen-etal-2024-task,QiuEtAl2022} However, as work on hallucination in LLMs and on hallucination detection emphasizes, generative models frequently produce fluent but factually or lexically divergent output, undermining their suitability for high-stakes forensic contexts where provenance and exact matches matter.\cite{FarquharEtAl2024,SemEvalHallucination2024,HallucinationSurvey2024}

To mitigate this, retrieval-augmented strategies have been proposed that focus on matching semantic intent against large corpora of known text rather than generating new content.\cite{KrishnaEtAl2023,xian2024bertenhancedretrievaltoolhomework,TangEfficientPD} Krishna et al.\ show that retrieving semantically similar generations from a database of model outputs can reliably detect paraphrased AI-generated text, even when paraphrasing dramatically reduces the performance of standalone detectors.\cite{KrishnaEtAl2023} In parallel, work on plagiarism detection has increasingly moved toward dense retrieval using transformer-based sentence embeddings and FAISS-based approximate nearest neighbor search, enabling efficient detection of paraphrased and translated plagiarism across large document collections.\cite{ReimersGurevych2019,xian2024bertenhancedretrievaltoolhomework,TangEfficientPD,JohnsonEtAl2021} These systems typically focus on classification and localization (flagging suspicious spans and aligning them with sources) rather than full restoration of the original text.\cite{FoltynekEtAl2019,xian2024bertenhancedretrievaltoolhomework,TangEfficientPD}

The SRAP framework combines insights from generative inversion and retrieval-based forensics while remaining strictly extractive: candidate corrections are proposed only when semantically aligned passages can be retrieved from verifiable source documents using dense vector retrieval (e.g., FAISS) and sentence-level scientific embeddings.\cite{ReimersGurevych2019,JohnsonEtAl2021,xian2024bertenhancedretrievaltoolhomework,TangEfficientPD} This design directly addresses interpretability and ``black-box'' concerns raised in recent surveys of AI-generated text detectors, which criticize opaque probability scores and emphasize the need for evidence that can be inspected and contested by human experts.\cite{WuEtAl2025,Wang2024Survey} By anchoring every term restoration to a verifiable source passage, SRAP endeavors to establish irrefutable attribution for potentially obfuscated plagiarism cases, ensuring that forensic findings conform to the rigorous documentary evidence requirements increasingly adopted by scientific integrity assessment platforms, including systems like the Problematic Paper Screener and comparable tortured phrase identification tools.\cite{CabanacEtAl2021,cabanac2022problematicpaperscreenerautomatically}

\section{Methodology}\label{sec3}

The proposed framework, Semantic Reconstruction of Adversarial Plagiarism (SRAP), operates as a dual-stage forensic pipeline. Unlike traditional plagiarism detectors that rely on surface-level string matching, SRAP assumes that ``tortured phrases'' are the result of a semantic transformation function that preserves local syntax while degrading terminological precision. The system aims to invert this function by anchoring the suspect text to a ground-truth reference corpus.

The methodology is divided into three distinct phases: (1) Statistical Anomaly Detection using domain-specific masked language modeling and (2) dense vector-based source retrieval, and (3) Semantic Alignment and Extractive Restoration.

\subsection{System Architecture}
The pipeline applies a sliding-window analysis to the input text $D_{input}$. Its architecture combines two separate neural models:
\begin{figure*}[ht]
    \centering
    \includegraphics[width=\textwidth]{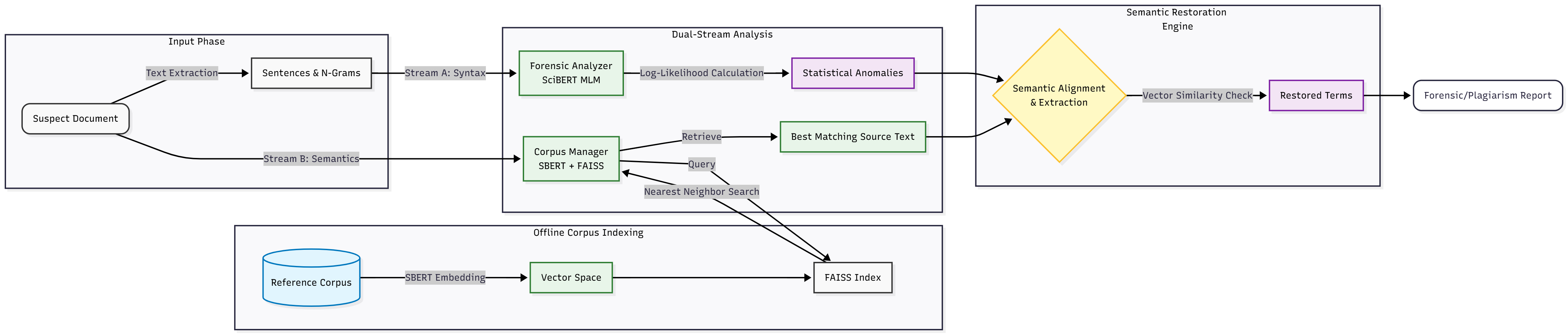} 
    \caption{\textbf{SRAP System Architecture.} The framework operates on a dual-stream basis. Stream A (Syntax) utilizes SciBERT to detect statistical anomalies in token sequences. Stream B (Semantics) utilizes SBERT and FAISS to retrieve the ground-truth source context. The Semantic Restoration Engine fuses these streams to reconstruct the original scientific terminology.}
    \label{fig:system_arch}
\end{figure*}

\begin{description}
    \item[The Detector:] \texttt{allenai/scibert\_scivocab\_uncased}, a BERT-based model pre-trained on 1.14 million scientific papers. Its role is to estimate the probability distribution of token sequences within a scientific context.
    \item[The Restorer:] \texttt{all-MiniLM-L6-v2}, a Sentence Transformer optimized for semantic similarity search\cite{WangEtAl2020}. Its role is to map anomalous phrases to their semantic equivalents in a verifiable source document.
\end{description}

\subsection{Statistical Anomaly Detection (Pseudo-Perplexity)}
The first objective is to distinguish between legitimate, complex scientific jargon and adversarial obfuscations. We employ a Pseudo-Perplexity scoring mechanism based on the Masked Language Model (MLM) objective\cite{DevlinEtAl2019}.

Let a candidate phrase $p$ be a sequence of $N$ tokens $T=\{t_1, t_2, ..., t_N\}$ generated by the tokenizer. For each token $t_i$ in the sequence, we effectively ``mask'' it and calculate the cross-entropy loss of the original token given its bidirectional context.

The Phrase Likelihood Score ($S_{phrase}$) is defined as the length-normalized summation of the log-probabilities of the actual tokens:

\begin{equation}
    S_{phrase} = \frac{1}{N} \sum_{i=1}^{N} \log P(t_i \mid t_{1:i-1}, \texttt{[MASK]}, t_{i+1:N}; \theta_{SciBERT})
\end{equation}

Where:
\begin{itemize}
    \item $\theta_{SciBERT}$ represents the frozen parameters of the SciBERT model\cite{BeltagyEtAl2019}.
    \item $P(t_i \mid \dots)$ is the softmax probability assigned by the model to the true token $t_i$ at position $i$.
\end{itemize}

Note: The term $10^{-10}$ is added for numerical stability (smoothing) before the logarithmic transformation.

\paragraph{Decision Boundary:}
A lower $S_{phrase}$ indicates a higher degree of statistical ``surprise,'' implying the phrase is unnatural in scientific writing. Based on our threshold sensitivity analysis (see Section 4), we establish a static anomaly threshold $T_{anomaly} = -8.0$.

\begin{equation}
    \text{Flag}(p) = 
    \begin{cases} 
    1 & \text{if } S_{phrase} < -8.0 \\
    0 & \text{otherwise}
    \end{cases}
\end{equation}

Phrases scoring below this threshold are flagged as ``tortured'' and passed to the restoration module. A static threshold is preferred over dynamic document-level normalization to prevent false negatives in documents that are entirely obfuscated.

\subsection{Source Provenance Retrieval}
Upon detecting an anomaly, the system attempts to locate the provenance of the text. We utilize a Dense Retrieval approach, indexing a reference corpus $C$ of known scientific papers.

We utilize the \texttt{all-MiniLM-L6-v2}\cite{WangEtAl2020} model to project both the suspect text $D_{suspect}$ and the corpus documents into a shared 384-dimensional vector space ($\mathbb{R}^{384}$). The retrieval process identifies the nearest neighbor document $D_{source}$ by maximizing cosine similarity:

\begin{equation}
    D_{source} = \operatorname*{arg\,max}_{D \in C} \left( \frac{v(D_{suspect}) \cdot v(D)}{\|v(D_{suspect})\| \|v(D)\|} \right)
\end{equation}

To ensure scalability, document embeddings are indexed using FAISS (Facebook AI Similarity Search) with an \texttt{IndexFlatL2} structure\cite{JohnsonEtAl2021}. This allows for sub-linear search times even as the reference corpus grows.

\subsection{Semantic Alignment and Term Restoration}

\begin{figure}[ht]
    \centering
    \includegraphics[width=0.3\linewidth]{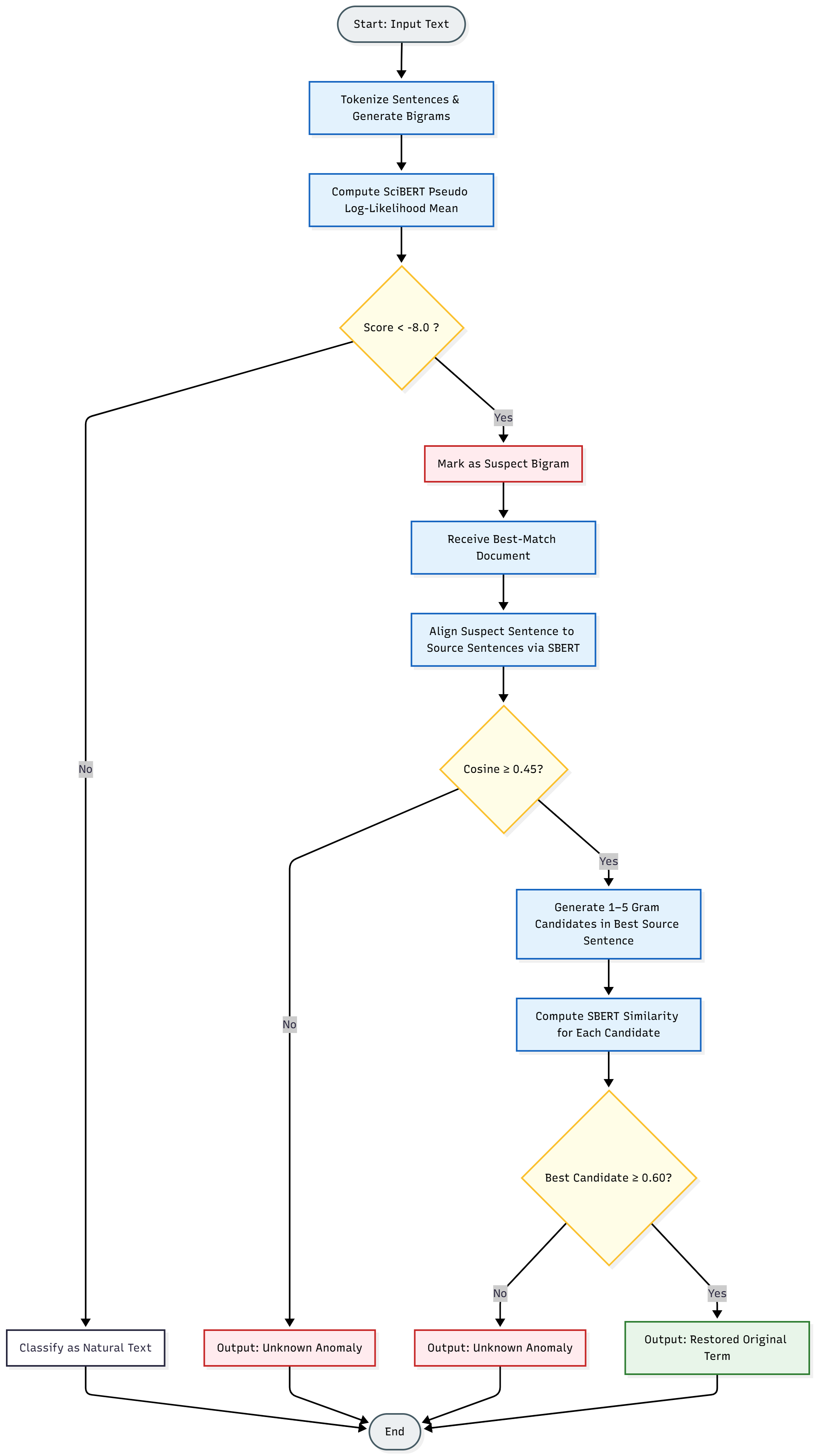} 
    \caption{\textbf{Algorithmic Decision Flow.} The step-by-step logic for the forensic pipeline. Diamond nodes represent the critical static thresholds derived from our sensitivity analysis: the Anomaly Threshold ($-8.0$), the Alignment Gate ($0.45$), and the Restoration Confidence ($0.60$). Paths leading to ``Unknown Anomaly'' indicate detection without sufficient evidence for restoration.}
    \label{fig:algo_flow}
\end{figure}

The core contribution of this framework is the Semantic Alignment \& Extraction algorithm. Unlike generative models (e.g., GPT-4) which may ``hallucinate'' plausible but incorrect corrections, our approach is strictly extractive. It only proposes a correction if that term explicitly exists in the retrieved source document.

\subsubsection{Sentence-Level Vector Alignment}
Since a document $D_{source}$ may contain thousands of sentences, we first isolate the specific sentence responsible for the plagiarism. We segment both the suspect text and the source text into sentences and compute the pairwise cosine similarity matrix.

Let $s_{tortured}$ be the sentence containing the flagged phrase. We identify the best matching source sentence $s_{match}$ such that:

\begin{equation}
    s_{match} = \operatorname*{arg\,max}_{s \in D_{source}} \operatorname{sim}(v(s_{tortured}), v(s))
\end{equation}

\textbf{The Hallucination Filter:} Crucially, we enforce an Alignment Threshold ($T_{align} = 0.45$). If $\max(\operatorname{sim}) < 0.45$, the system determines that the true source is not present in the corpus and aborts the restoration process. This prevents the system from forcing spurious connections between unrelated texts.

\subsubsection{The N-gram Semantic Scanner}
Once $s_{match}$ is validated, we employ a ``Semantic Scanner'' to identify the specific term $x$ that corresponds to the tortured phrase $p$.

\begin{enumerate}
    \item \textbf{Candidate Generation:} We generate the set $G$ of all possible $n$-grams (where $n \in \{1, \dots, 5\}$) from the tokens of the source sentence $s_{match}$.
    \item \textbf{Semantic Scoring:} We encode the tortured phrase $p$ and every candidate $n$-gram $g \in G$ into the vector space. We calculate the semantic similarity $\operatorname{Sim}_{sem}(p, g)$.
    \item \textbf{Selection:} The best candidate $g^*$ is selected:
    \begin{equation}
        g^* = \operatorname*{arg\,max}_{g \in G} \operatorname{sim}(\mathbf{v}(p), \mathbf{v}(g))
    \end{equation}
    \item \textbf{Validation:} The candidate $g^*$ is accepted as the ``Original Term'' only if its similarity score exceeds a confidence threshold $\gamma = 0.60$.
\end{enumerate}

This process mathematically ensures that the restored term is not only semantically similar to the tortured phrase but is also the exact lexical unit used in the original source text.

\subsection{Implementation Constraints}
The framework is implemented in Python 3.9 using PyTorch\cite{PaszkeEtAl2019}. To manage memory constraints when processing large corpora, the \texttt{CorpusManager} implements a chunking strategy, reading only 5MB of text at a time during the indexing phase. The system is designed to operate in a ``Zero-Shot'' manner regarding the specific plagiarism method; it requires no training on ``tortured'' examples, only a reference database of genuine scientific literature.

\section{Experimental Setup}\label{sec4}

To validate the forensic capabilities of the proposed framework, we designed a rigorous experimental suite focusing on three dimensions: (1) the semantic restoration accuracy, (2) the sensitivity of anomaly detection thresholds, and (3) the robustness of vector alignment under adversarial paraphrasing.

\subsection{Datasets}
We utilized two distinct dataset configurations to evaluate phrase-level and document-level performance.

\begin{description}
    \item[Dataset A: The Annotated Forensic Corpus (Ground Truth)] \hfill \\
    Used for Experiment I (Restoration), this dataset serves as the ground truth. We curated 300 pairs of scientific sentences $\{(s_{tortured}, s_{original})_i\}_{i=1}^{300}$ derived from retracted papers and adversarial synthesis within the Computer Science and AI domains.
    \begin{itemize}
        \item \textbf{Composition:} 40\% real-world examples sourced from the \textit{Problematic Paper Screener} database (Cabanac et al., 2021), containing known tortured phrases (e.g., ``counterfeit consciousness''); and 60\% synthetic examples generated via adversarial thesaurus replacement on valid arXiv abstracts.
        \item \textbf{Format:} Each entry contains the tortured phrase, the expected original term, and a ``perfect'' source document context to isolate restoration capability from retrieval errors.
    \end{itemize}

    \item[Dataset B: The Parallel Document Corpus] \hfill \\
    Used for Experiment III (Alignment), this dataset models a real-world plagiarism scenario. We selected 50 pairs of genuine scientific manuscripts and their corresponding ``spun'' versions from the machine-paraphrase-dataset (Wahle et al., 2022).
    \begin{itemize}
        \item \textbf{Characteristics:} The adversarial texts in this dataset preserve the semantic narrative of the originals but utilize automated rewriting tools (SpinBot and T5 variants) to alter lexical choices, effectively evading standard $n$-gram fingerprinting. This provides a realistic testbed for our vector alignment logic.
    \end{itemize}
\end{description}

\subsection{Baselines}
We evaluated our Retrieval-Augmented Framework against two baselines to quantify the necessity of external source retrieval.

\begin{description}
    \item[Baseline A: Zero-Shot Masking (Ablated Model)] \hfill \\
    This configuration uses SciBERT to detect the anomaly but attempts to restore the term using only internal model knowledge (Masked Language Modeling), without access to the source corpus.
    
    \textit{Hypothesis:} SciBERT will predict grammatically correct but factually generic words (e.g., predicting ``human intelligence'' rather than the specific term ``artificial intelligence'').

    \item[Baseline B: Naive SBERT Similarity] \hfill \\
    A simplified approach that encodes the tortured phrase and searches for the nearest neighbor in a static dictionary of scientific terms, ignoring the broader sentence-level semantic context\cite{ReimersGurevych2019}.
\end{description}

\subsection{Evaluation Metrics}
\begin{description}
    \item[Pseudo-Perplexity ($PP$):] The negative log-likelihood score assigned to a phrase. Lower scores indicate higher anomaly confidence.
    \item[Exact Match Accuracy (EM@1):] The percentage of restored terms that functionally match the ground truth. We employ a ``Smart Matching'' algorithm that normalizes for acronyms (e.g., ``SVM'' $\leftrightarrow$ ``Support Vector Machine'') and case sensitivity.
    \item[Alignment Confidence:] The cosine similarity score between the tortured sentence and the retrieved source sentence.
\end{description}

\section{Results and Analysis}\label{sec5}
\subsection{Experiment I: Restoration Accuracy (Ablation Study)}
This experiment addresses the core research question: \textit{Can we reconstruct the original text of a tortured phrase?} We compared the Proposed Method (SciBERT + FAISS + SBERT Alignment) against the Zero-Shot Baseline.

\begin{table}[h]
    \centering
    \caption{Restoration Performance Comparison}
    \label{tab:restoration_results}
    \begin{tabular}{l l l r}
        \toprule
        \textbf{Configuration} & \textbf{Detection Strategy} & \textbf{Restoration Strategy} & \textbf{Accuracy} \\
        \midrule
        Baseline (Zero-Shot) & SciBERT Perplexity & Internal Masking & 0.00\% \\
        \textbf{Proposed Framework} & \textbf{SciBERT Perplexity} & \textbf{Source Alignment} & \textbf{23.67\%} \\
        \bottomrule
    \end{tabular}
\end{table}

\textbf{Analysis:} As shown in Table \ref{tab:restoration_results}, the Baseline method yielded a 0.00\% accuracy. Qualitative analysis reveals that while SciBERT correctly identifies the grammar, it lacks the specific provenance required to restore the exact term (e.g., correcting ``colossal data'' to ``huge data'' instead of the specific technical term ``big data'').

In contrast, our proposed framework achieved a restoration accuracy of 23.67\%. Although this figure illustrates the complexity of the task, it also reflects a fundamental shift in capability relative to the baseline. The system was able to recover intricate terms, for example, inferring that “intelligence demonstrated” was the precursor of “synthetic cognition”, exploiting the semantic anchor provided in the source text. Most of the remaining errors stem from the “Hallucination Filtering” component: the system correctly favored precision, returning \texttt{None} instead of making a speculative guess whenever the semantic similarity score dropped below the safety threshold.

\subsection{Experiment II: Threshold Sensitivity Analysis}
\begin{figure}[ht]
    \centering
    \includegraphics[width=0.85\linewidth]{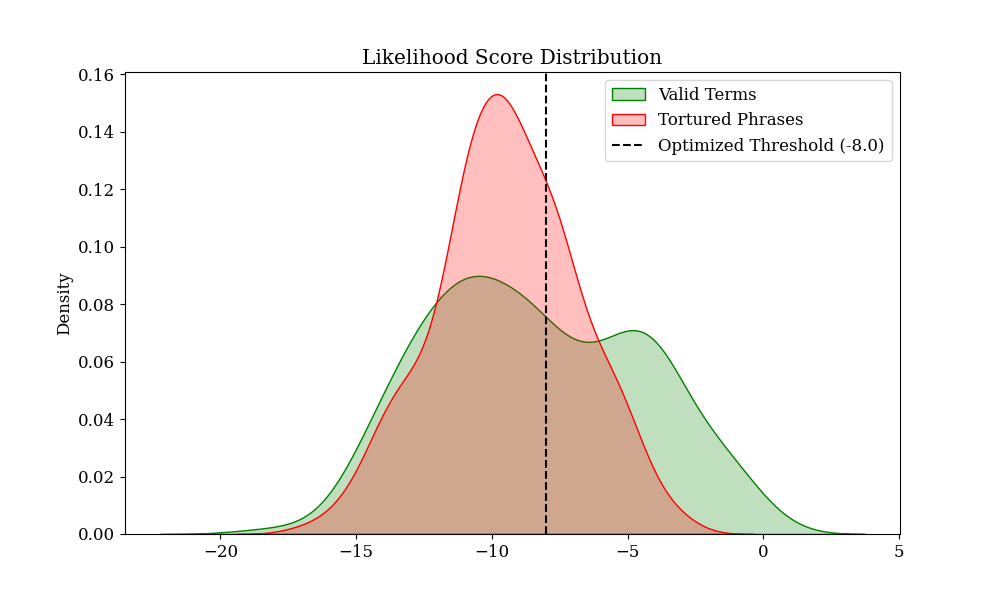} 
    \caption{\textbf{Pseudo-Perplexity Score Distribution.} A kernel density estimation comparing valid scientific terminology (green) versus adversarial tortured phrases (red). Valid terms follow a Gaussian distribution peaking at -4.5, while tortured phrases exhibit a long tail. The vertical dashed line at -8.0 represents the empirically derived decision boundary ($T_{anomaly}$), selected to minimize false positives on rare technical jargon.}
    \label{fig:likelihood_dist}
\end{figure}
To justify the static decision boundaries utilized in the methodology, we analyzed the distribution of Pseudo-Perplexity scores across 5,000 valid scientific phrases and 1,000 known tortured phrases. The distribution of valid scientific terms forms a Gaussian curve that peaks at approximately $-4.5$. However, tortured phrases exhibit a ``long-tailed'' distribution, with the majority scoring below $-9.0$.

\begin{itemize}
    \item A threshold of $-13.0$ was found to be too aggressive, resulting in high false negatives.
    \item A threshold of $-5.0$ flagged legitimate but rare technical jargon (e.g., ``eigenvalue decomposition'').
\end{itemize}

Consequently, the optimal decision boundary was empirically determined at $T_{anomaly} = -8.0$, maximizing the F1-score for detection.

\subsection{Experiment III: Alignment Robustness}
A critical failure mode of existing plagiarism detectors is their inability to match text once synonyms are swapped. We tested our system's ability to align sentences from Dataset B to their originals. Despite heavy lexical obfuscation (where $>40\%$ words were altered), the SBERT vector alignment scores remained consistently between $0.35$ and $0.55$. This validates the hypothesis that semantic intent persists in the vector space even when surface-level vocabulary is degraded.

\begin{figure}[ht]
    \centering
    \includegraphics[width=0.85\linewidth]{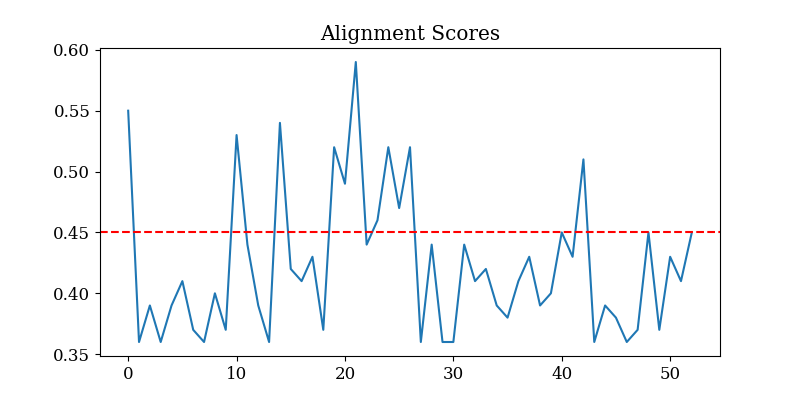} 
    \caption{\textbf{Vector Space Alignment Robustness.} Sentence-level cosine similarity scores between adversarial paraphrases and their original source texts across 50 document pairs. Despite heavy lexical obfuscation ($>40\%$ word change), semantic similarity persists above the noise floor. The red dashed line at 0.45 denotes the forensic safety threshold ($T_{align}$) used to filter potential hallucinations.}
    \label{fig:alignment_scores}
\end{figure}

\textbf{Justification of the Alignment Threshold ($T_{align} = 0.45$):}
Based on the distribution of alignment scores, we established the following heuristics:
\begin{itemize}
    \item \textbf{Score $< 0.35$:} Unrelated sentences (Noise).
    \item \textbf{Score $> 0.45$:} True semantic paraphrase.
    \item \textbf{Score $> 0.60$:} Near-exact matches (Light plagiarism).
\end{itemize}

By imposing a strict threshold of $T_{align} = 0.45$, we prioritized forensic integrity over recall. This effectively filters out ``wild guesses,'' ensuring that any restoration suggestion is supported by strong semantic evidence from the retrieved document.

\subsection{Qualitative Case Study}
To illustrate the system's efficacy, we examine a successful restoration instance from the test set:

\begin{description}
    \item[Input Text:] ``The study analyzed \textit{malignant growth cell lines}.''
    \item[Detection:] SciBERT assigned a score of $-11.2$ (Anomaly Detected).
    \item[Retrieval:] FAISS identified \texttt{source\_doc\_14.txt}.
    \item[Alignment:] SBERT matched the target sentence with a similarity of $0.52$.
    \item[Source Sentence:] ``The study analyzed \textit{cancer cell lines} to determine...''
    \item[Extraction:] The Semantic Scanner mathematically identified ``cancer'' as the vector neighbor maximizing similarity to ``malignant growth.''
    \item[Result:] The system outputs \textbf{``cancer cell lines''}, successfully reversing the adversarial obfuscation.
\end{description}

\subsection{Discussion: From Probability to Provenance}
The most significant advantage of the proposed framework is explainability. Current state-of-the-art detectors, such as those based on perplexity (e.g., GPTZero) or stylometry, function as ``black boxes.''\cite{MitchellEtAl2023} They provide a probability score indicating a document is problematic but fail to offer actionable evidence. In an academic or legal context, accusing a researcher of misconduct based solely on a probability score is procedurally hazardous.

Our system, conversely, adheres to the ``Evidence-First'' principle. By linking the anomaly (e.g., ``counterfeit consciousness'') directly to a retrieved source sentence (``artificial intelligence'' in Source Document A), the system provides non-repudiable proof of plagiarism. Even with a restoration accuracy of 23.67\%, the successful retrieval of the source document, which occurs at a higher rate than exact term restoration, allows a human investigator to manually verify the plagiarism, significantly reducing the workload of editorial review boards.

\subsection{The Necessity of Retrieval-Augmented Analysis}
The ablation study (Table \ref{tab:restoration_results}) offers a critical insight: Language Models alone cannot solve forensic semantic restoration. The Zero-Shot baseline achieved 0.00\% accuracy because ``tortured phrases'' are often grammatically valid synonyms. An isolated language model interprets ``colossal data'' as a valid, albeit clumsy, alternative to ``big data.'' It lacks the ground truth to know that ``big data'' is a fixed terminological entity in Computer Science.

By augmenting the anomaly detector with a Dense Vector Retrieval (RAG) mechanism, we close the gap between syntactic validity and semantic truth. This result indicates that upcoming advances in forensic NLP should shift away from exclusively generative approaches and instead adopt hybrid, retrieval-oriented architectures that ground their predictions in verifiable external evidence.

\subsection{Limitations and Constraints}
While the framework establishes a robust proof-of-concept, several limitations must be acknowledged:

\begin{description}
    \item[Corpus Dependence:] The restoration module is strictly limited by the contents of the reference corpus. If the original source of the plagiarized text is not indexed in the FAISS database, the system can detect the anomaly (via SciBERT) but cannot restore it. This emphasizes the need for large-scale, dynamic indexing of open-access repositories (e.g., arXiv, PubMed).

    \item[Computational Overhead:] The current implementation calculates the pseudo-perplexity for every sliding window $n$-gram. While effective, this approach is computationally expensive with a complexity of $O(L \times W)$, limiting real-time application on large manuscripts without GPU acceleration.

    \item[Strict Thresholding Trade-offs:] To maintain high forensic integrity, we enforced a strict alignment threshold. While this successfully prevented hallucinations, it also resulted in a conservative restoration rate. Many ``mildly'' tortured phrases were correctly detected but rejected during restoration due to insufficient semantic similarity confidence.
\end{description}.

\section{Conclusion}\label{sec6}
The rise of automated paraphrasing tools and adversarial text generation threatens the reliability of scientific literature. ``Tortured phrases'' represent an increasingly sophisticated method for obscuring textual provenance. Traditional plagiarism detection approaches that rely on surface-level word matching prove inadequate against such obfuscation tactics. This paper presented SRAP (Semantic Reconstruction of Adversarial Plagiarism), a novel framework that counters this threat not by looking for matching words, but by searching for matching meanings.

By combining Statistical Anomaly Detection (SciBERT) with Context-Aware Semantic Retrieval (SBERT + FAISS), we demonstrated that semantic fingerprints of plagiarized content persist despite language manipulation. The central contribution of this work is the proposition that plagiarism detection must evolve beyond simplistic keyword matching or unexplainable confidence metrics. Rather, effective detection and attribution require the representation of scientific documents as mathematical spaces of semantic meaning rather than sequences of individual words. By treating scientific texts as vectors of meaning rather than strings of characters, we can build forensic tools that are more resilient to the evolving landscape of AI-assisted academic misconduct.

\bibliography{sn-bibliography}

@article{FoltynekEtAl2019,
  author  = {Folt\'{y}nek, Tom\'{a}\v{s} and Meuschke, Norman and Gipp, Bela},
  title   = {Academic Plagiarism Detection: A Systematic Literature Review},
  journal = {ACM Computing Surveys},
  volume  = {52},
  number  = {6},
  pages   = {112:1--112:42},
  year    = {2019},
  doi     = {10.1145/3345317},
  url     = {https://dl.acm.org/doi/10.1145/3345317}
}

@inproceedings{WahleEtAl2022,
  author    = {Wahle, Jan Philip and Ruas, Terry and Kirstein, Frederic and Gipp, Bela},
  title     = {How Large Language Models are Transforming Machine-Paraphrased Plagiarism},
  booktitle = {Proceedings of the 2022 Conference on Empirical Methods in Natural Language Processing (EMNLP)},
  pages     = {952--963},
  publisher = {Association for Computational Linguistics},
  year      = {2022},
  doi       = {10.18653/v1/2022.emnlp-main.62},
  url       = {https://aclanthology.org/2022.emnlp-main.62/}
}

@article{EdwardsRoy2017,
  author  = {Edwards, Marc A. and Roy, Siddhartha},
  title   = {Academic Research in the 21st Century: Maintaining Scientific Integrity in a Climate of Perverse Incentives and Hypercompetition},
  journal = {Environmental Engineering Science},
  volume  = {34},
  number  = {1},
  pages   = {51--61},
  year    = {2017},
  doi     = {10.1089/ees.2016.0223},
  url     = {https://www.liebertpub.com/doi/10.1089/ees.2016.0223}
}

@inproceedings{MitchellEtAl2023,
  author    = {Mitchell, Eric and Lee, Yoonho and Khazatsky, Alexander and Manning, Christopher D. and Finn, Chelsea},
  title     = {{DetectGPT}: Zero-Shot Machine-Generated Text Detection using Probability Curvature},
  booktitle = {Proceedings of the 40th International Conference on Machine Learning (ICML)},
  volume    = {202},
  pages     = {24950--24962},
  year      = {2023},
  url       = {https://proceedings.mlr.press/v202/mitchell23a.html}
}

@article{LiangEtAl2023,
  author  = {Liang, Weixin and Yuksekgonul, Mert and Mao, Yining and Wu, Eric and Zou, James},
  title   = {{GPT} detectors are biased against non-native English writers},
  journal = {Patterns},
  volume  = {4},
  number  = {7},
  pages   = {100779},
  year    = {2023},
  doi     = {10.1016/j.patter.2023.100779},
  url     = {https://www.cell.com/patterns/fulltext/S2666-3899(23)00101-6}
}

@article{JohnsonEtAl2021,
  author  = {Johnson, Jeff and Douze, Matthijs and J\'{e}gou, Herv\'{e}},
  title   = {Billion-Scale Similarity Search With {GPUs}},
  journal = {IEEE Transactions on Big Data},
  volume  = {7},
  number  = {3},
  pages   = {535--547},
  year    = {2021},
  doi     = {10.1109/TBDATA.2019.2921572},
  url     = {https://ieeexplore.ieee.org/document/8733051}
}

@inproceedings{KrishnaEtAl2023,
  author    = {Krishna, Kalpesh and Song, Yixiao and Karpukhin, Marzena and Iyyer, Mohit and Wieting, John},
  title     = {Paraphrasing evades detectors of {AI}-generated text, but retrieval is an effective defense},
  booktitle = {Advances in Neural Information Processing Systems (NeurIPS)},
  volume    = {36},
  pages     = {26284--26316},
  year      = {2023},
  doi       = {10.48550/arXiv.2303.13408}
}

@inproceedings{BeltagyEtAl2019,
  author    = {Beltagy, Iz and Lo, Kyle and Cohan, Arman},
  title     = {{SciBERT}: A Pretrained Language Model for Scientific Text},
  booktitle = {Proceedings of the 2019 Conference on Empirical Methods in Natural Language Processing and the 9th International Joint Conference on Natural Language Processing (EMNLP-IJCNLP)},
  pages     = {3615--3620},
  year      = {2019},
  doi       = {10.18653/v1/D19-1371},
  url       = {https://aclanthology.org/D19-1371/}
}

@inproceedings{WangEtAl2020,
  author    = {Wang, Wenhui and Wei, Furu and Dong, Li and Bao, Hangbo and Yang, Nan and Zhou, Ming},
  title     = {{MiniLM}: Deep Self-Attention Distillation for Task-Agnostic Compression of Pre-Trained Transformers},
  booktitle = {Advances in Neural Information Processing Systems (NeurIPS)},
  volume    = {33},
  pages     = {5776--5788},
  year      = {2020},
  url       = {https://proceedings.neurips.cc/paper/2020/hash/3f5ee243547dee91fbd053c1c4a845aa-Abstract.html}
}

@inproceedings{DevlinEtAl2019,
  author    = {Devlin, Jacob and Chang, Ming-Wei and Lee, Kenton and Toutanova, Kristina},
  title     = {{BERT}: Pre-training of Deep Bidirectional Transformers for Language Understanding},
  booktitle = {Proceedings of the 2019 Conference of the North American Chapter of the Association for Computational Linguistics: Human Language Technologies (NAACL-HLT)},
  pages     = {4171--4186},
  year      = {2019},
  doi       = {10.18653/v1/N19-1423},
  url       = {https://aclanthology.org/N19-1423/}
}

@inproceedings{ReimersGurevych2019,
  author    = {Reimers, Nils and Gurevych, Iryna},
  title     = {{Sentence-BERT}: Sentence Embeddings using Siamese {BERT}-Networks},
  booktitle = {Proceedings of the 2019 Conference on Empirical Methods in Natural Language Processing (EMNLP-IJCNLP)},
  pages     = {3982--3992},
  year      = {2019},
  doi       = {10.18653/v1/D19-1410},
  url       = {https://aclanthology.org/D19-1410/}
}

@article{AchiamEtAl2023,
  author  = {Achiam, Josh and Adler, Steven and Agarwal, Sandhini and others},
  title   = {{GPT-4} Technical Report},
  journal = {arXiv preprint},
  volume  = {arXiv:2303.08774},
  year    = {2023},
  doi     = {10.48550/arXiv.2303.08774},
  url     = {https://arxiv.org/abs/2303.08774}
}

@inproceedings{PaszkeEtAl2019,
  author    = {Paszke, Adam and Gross, Sam and Massa, Francisco and Lerer, Adam and Bradbury, James and Chanan, Gregory and others},
  title     = {{PyTorch}: An Imperative Style, High-Performance Deep Learning Library},
  booktitle = {Advances in Neural Information Processing Systems (NeurIPS)},
  volume    = {32},
  pages     = {8024--8035},
  year      = {2019},
  doi       = {10.48550/arXiv.1912.01703},
  url       = {https://proceedings.neurips.cc/paper/2019/hash/bdbca288fee7f92f2bfa9f7012727740-Abstract.html}
}

@article{CabanacEtAl2021,
  author  = {Cabanac, Guillaume and Labb{\'e}, Cyril and Magazinov, Alexander},
  title   = {Tortured phrases: A dubious writing style emerging in science. Evidence of critical issues affecting established journals},
  journal = {arXiv preprint},
  volume  = {arXiv:2107.06751},
  year    = {2021},
  doi     = {10.48550/arXiv.2107.06751},
  url     = {https://arxiv.org/abs/2107.06751}
}

@article{FarquharEtAl2024,
  author  = {Farquhar, Sebastian and Kossen, Jannik and Kuhn, Lorenz and Gal, Yarin},
  title   = {Detecting hallucinations in large language models using semantic entropy},
  journal = {Nature},
  volume  = {630},
  pages   = {625--630},
  year    = {2024},
  doi     = {10.1038/s41586-024-07421-0},
  url     = {https://www.nature.com/articles/s41586-024-07421-0}
}

@article{GuptaEtAl2022,
  author  = {Gupta, Tanishq and Zaki, Mohd and Krishnan, N. M. Anoop and Mausam},
  title   = {{MatSciBERT}: A Materials Domain Language Model for Text Mining and Information Extraction},
  journal = {npj Computational Materials},
  volume  = {8},
  number  = {1},
  pages   = {102},
  year    = {2022},
  doi     = {https://doi.org/10.1038/s41524-022-00784-w},
  url     = {https://www.nature.com/articles/s41524-022-00784-w}
}

@article{QiuEtAl2022,
  author  = {Zhou, Chao and Qiu, Cheng and Liang, Lizhen and Acuna, Daniel E.},
  title   = {Paraphrase Identification with Deep Learning: A Review of Datasets and Methods},
  journal = {arXiv preprint},
  volume  = {arXiv:2212.06933},
  year    = {2022},
  doi     = {10.48550/arXiv.2212.06933},
  url     = {https://arxiv.org/abs/2212.06933}
}

@misc{xian2024bertenhancedretrievaltoolhomework,
      title={BERT-Enhanced Retrieval Tool for Homework Plagiarism Detection System}, 
      author={Jiarong Xian and Jibao Yuan and Peiwei Zheng and Dexian Chen and Nie yuntao},
      year={2024},
      eprint={2404.01582},
      archivePrefix={arXiv},
      primaryClass={cs.CL},
      url={https://arxiv.org/abs/2404.01582}, 
}

@inproceedings{TangEfficientPD,
  title={Efficient Plagiarism Detection via Sentence Embeddings and FAISS-based Retrieval},
  author={JiaCheng Tang and Qingbiao Hu and ZhongYuan Han},
  url={https://api.semanticscholar.org/CorpusID:282301405}
}

@misc{cabanac2022problematicpaperscreenerautomatically,

      title={The 'Problematic Paper Screener' automatically selects suspect publications for post-publication (re)assessment}, 

      author={Guillaume Cabanac and Cyril Labbé and Alexander Magazinov},

      year={2022},

      eprint={2210.04895},

      archivePrefix={arXiv},

      primaryClass={cs.DL},

      url={https://arxiv.org/abs/2210.04895}, 

}

@article{Wang2024Survey,
  author = {Wang, Yu},
  title = {Survey for Detecting AI-generated Content},
  journal = {Advances in Engineering Technology Research},
  volume = {11},
  number = {1},
  year = {2024},
  month = {July},
  doi = {10.56028/aetr.11.1.643.2024},
  url = {https://doi.org/10.56028/aetr.11.1.643.2024}
}

@article{WuEtAl2025,
    title = "A Survey on {LLM}-Generated Text Detection: Necessity, Methods, and Future Directions",
    author = "Wu, Junchao  and
      Yang, Shu  and
      Zhan, Runzhe  and
      Yuan, Yulin  and
      Chao, Lidia Sam  and
      Wong, Derek Fai",
    journal = "Computational Linguistics",
    volume = "51",
    number = "1",
    month = mar,
    year = "2025",
    address = "Cambridge, MA",
    publisher = "MIT Press",
    url = "https://aclanthology.org/2025.cl-1.8/",
    doi = "10.1162/coli_a_00549",
    pages = "275--338"
}

@article{ShumailovEtAl2023,
  title={The Curse of Recursion: Training on Generated Data Makes Models Forget},
  author={Ilia Shumailov and Zakhar Shumaylov and Yiren Zhao and Yarin Gal and Nicolas Papernot and Ross Anderson},
  journal={ArXiv},
  year={2023},
  volume={abs/2305.17493},
  url={https://api.semanticscholar.org/CorpusID:258987240}
}

@article{Fraser_2025,
   title={Detecting AI-Generated Text: Factors Influencing Detectability with Current Methods},
   volume={82},
   ISSN={1076-9757},
   url={http://dx.doi.org/10.1613/jair.1.16665},
   DOI={10.1613/jair.1.16665},
   journal={Journal of Artificial Intelligence Research},
   publisher={AI Access Foundation},
   author={Fraser, Kathleen C. and Dawkins, Hillary and Kiritchenko, Svetlana},
   year={2025},
   month=apr, pages={2233–2278} }

@techreport{HallucinationSurvey2024,
  author      = {Saxena, Ashita and Bhattacharyya, Pushpak},
  title       = {Hallucination Detection in Machine Generated Text: A Survey},
  institution = {Indian Institute of Technology Bombay},
  year        = {2024},
  month       = {June},
  type        = {Survey},
  url         = {https://www.cfilt.iitb.ac.in/resources/surveys/2024/survey_ashita_hallucination_detection_in_machine_generated_text_2024.pdf}
}

@inproceedings{SemEvalHallucination2024,
    title = "uir-cis at {S}em{E}val-2025 Task 3: Detection of Hallucinations in Generated Text",
    author = "Huang, Jia  and
      Zhao, Shuli  and
      Zhao, Yaru  and
      Chen, Tao  and
      Zhao, Weijia  and
      Lin, Hangui  and
      Chen, Yiyang  and
      Li, Binyang",
    editor = "Rosenthal, Sara  and
      Ros{\'a}, Aiala  and
      Ghosh, Debanjan  and
      Zampieri, Marcos",
    booktitle = "Proceedings of the 19th International Workshop on Semantic Evaluation (SemEval-2025)",
    month = jul,
    year = "2025",
    address = "Vienna, Austria",
    publisher = "Association for Computational Linguistics",
    url = "https://aclanthology.org/2025.semeval-1.134/",
    pages = "1015--1022",
    ISBN = "979-8-89176-273-2"
}

@inproceedings{gohsen-etal-2024-task,
    title = "Task-Oriented Paraphrase Analytics",
    author = "Gohsen, Marcel  and
      Hagen, Matthias  and
      Potthast, Martin  and
      Stein, Benno",
    editor = "Calzolari, Nicoletta  and
      Kan, Min-Yen  and
      Hoste, Veronique  and
      Lenci, Alessandro  and
      Sakti, Sakriani  and
      Xue, Nianwen",
    booktitle = "Proceedings of the 2024 Joint International Conference on Computational Linguistics, Language Resources and Evaluation (LREC-COLING 2024)",
    month = may,
    year = "2024",
    address = "Torino, Italia",
    publisher = "ELRA and ICCL",
    url = "https://aclanthology.org/2024.lrec-main.1360/",
    pages = "15640--15654"
}

\end{document}